\documentclass[letterpaper]{article} 
\usepackage{aaai24}  
\usepackage{times}  
\usepackage{helvet}  
\usepackage{courier}  
\usepackage[hyphens]{url}  
\usepackage{graphicx} 
\usepackage{natbib}  
\usepackage{caption} 
\usepackage{algorithm}
\usepackage{algorithmic}
\usepackage{xcolor}

\usepackage{amsmath}
\usepackage{amssymb}
\usepackage{amsthm}
\DeclareMathOperator{\E}{\mathbb{E}}
\usepackage{newfloat}
\usepackage{listings}
\DeclareCaptionStyle{ruled}{labelfont=normalfont,labelsep=colon,strut=off} 
\floatstyle{ruled}
\newfloat{listing}{tb}{lst}{}
\floatname{listing}{Listing}
\nocopyright
\title{Closed Drafting as a Case Study for First-Principle Interpretability, Memory, and Generalizability in Deep Reinforcement Learning}
\author {
    Ryan Rezai\textsuperscript{\rm 1,}\equalcontrib,
    Jason Wang\textsuperscript{\rm 2,}\equalcontrib
}
\affiliations {
    \textsuperscript{\rm 1}University of Waterloo,
    \textsuperscript{\rm 2}Harvard University\\
    rrezai@uwaterloo.ca, jasonwang1@college.harvard.edu
}

\begin{document}

\maketitle

\begin{abstract}
Closed drafting or ``pick and pass" is a popular game mechanic where each round players select a card or other playable element from their hand and pass the rest to the next player. In this paper, we establish first-principle methods for studying the interpretability, generalizability, and memory of  Deep Q-Network (DQN) models playing closed drafting games. In particular, we use a popular family of closed drafting games called ``Sushi Go Party", in which we achieve state-of-the-art performance. We fit decision rules to interpret the decision-making strategy of trained DRL agents by comparing them to the ranking preferences of different types of human players. As Sushi Go Party can be expressed as a set of closely-related games based on the set of cards in play, we quantify the generalizability of DRL models trained on various sets of cards, establishing a method to benchmark agent performance as a function of environment unfamiliarity. Using the explicitly calculable memory of other player's hands in closed drafting games, we create measures of the ability of DRL models to learn memory.

\end{abstract}

\section{Introduction}


The field of deep reinforcement learning (DRL) is experiencing promising advancement in solving a variety of difficult problems \cite{li2018deep}. At the same time, several issues have been identified that prevent widespread deployment of DRL real-world tasks. One of the most commonly cited issues is a lack of agent interpretability \cite{glanois2021interpretable}. The policies developed by DRL models are often not able to be understood by humans. This presents problems in troubleshooting, regulation, and oversight. A related problem is in the ability of DRL models to perform reliably in unseen environments \cite{zhange2018overfitting}. DRL is often used for tasks that have partial observablility, for which memory is a crucial factor. Despite this, the ability of a human to interpret the role of memory in current DRL approaches is limited \cite{paischer2023semantic}.

There are elements of closed drafting games that are useful for studying interpretability, generalizability, and memory in DRL. Interpreting model decision-making is made easier by the limited number of cards that are available for the model to choose from at a given time. This means that general preferences for one card over another can be measured, and the learned decision-making strategy of models can be summarized in a way that is understandable to laypeople. Generalizability can be studied through the highly customizable nature of card drafting, where different sets of cards can be in play, and each configuration is related to another in measurable degrees. Memory can be studied due to the partially observability present in closed drafting games, and due to fact that memory of past hands is part of successful strategies, but is not directly provided to players. This means that we can explicitly include or exclude memory to study how/if agents learn memory.

To our knowledge, no prior works have studied closed drafting, although a few have incidentally used game environments with the mechanic. In particular, the bestseller Sushi Go was first studied by \cite{soen2019making} who simplified the setting to a full-information one, \cite{liu2020reinforcementlearning} who introduced a baseline of a rule-based agent following a set card priority ranking, and \cite{klang2021assessing} who placed Sushi Go in the general setting of tabletop games, although without the specificity of closed drafting we intend to study.

Our contributions in this paper are to (1) create a closed drafting Sushi Go Party multi-agent DRL environment that supports the explicit inclusion/exclusion of memory, (2) establish metrics for understanding DRL behavior across memory conditions, (3) quantify generalization dynamics to identify the rate of performance decrease out-of-distribution, and (4) demonstrate a more penetrating ability to interpret model decision-making with decision rules and preference rankings made possible by the closed drafting framework.

\begin{figure}[H]
    \centering
    \includegraphics[scale=0.45]{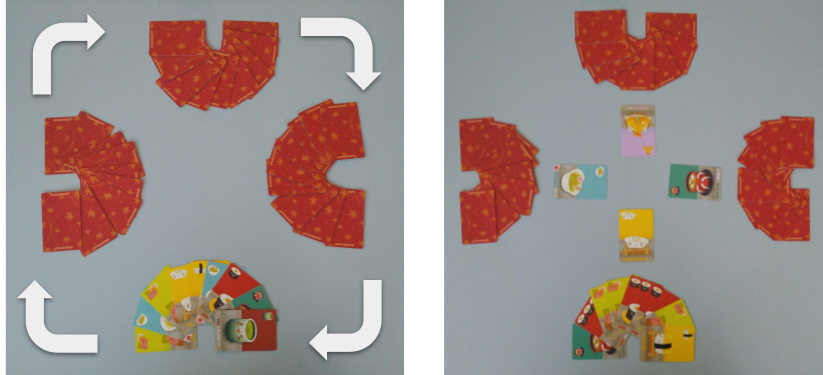}
    \caption{An Example of Initial Dealt Hands and First Turn for a Four-Player Sushi Go Party Game using the ``My First Meal" Configuration}
    \label{fig:hands}
\end{figure}

\paragraph{Setup} 

Sushi Go Party is played in groups of 2 to 8 players. Our experiments will be in groups of 4 players. Hands of 9 cards are dealt to each player. These hands are concealed, and not seen by the other players. Each player picks a single card, places it face down on the board, then delivers the remaining hand to a person next to them. After this the turn ends and they flip over and reveal the card they placed face down. This card remains face up on the board until the round is done. This process repeats until all cards are played, which will end the round. Due to the inability to explicitly see the hands of other players, this process is partially observable. Points are derived from the cards that are played on the board, which often have interactions (for example, the ``Tempura" card only yields points if two are in play). This process repeats for 3 rounds, and the winner is the player with the most points in the end.

We define closed drafting games generally as finite-horizon POMDPs $(\mu,S,A,P,r,H)$: the tuple of initial state distribution, state space, action space, state transitions, reward function, and horizon length. We encode a hand of cards as a frequency vector $(c_1,c_2,\dots,c_n)^T$ where $c_i$ denotes the number of cards of kind $i$ in the hand, and $n$ denotes the total number of unique cards. The action space is the kind of card to play $A=\{1,\dots,n\}$. The state space is specified by the hands of each player and the played cards, but the observation excludes the hands of the other players. Closed drafting mechanics introduce the restriction that the length of each hand is equal, and the length of a hand and the length of a player's played cards sums to $H$. Further, the transition dynamic is a simple function that simply rotates the hands---this enables us to work out some of the cards of other hands, and we can optionally explicitly provide the optimally sleuthed information in the observation. It is this simple transition dynamic that makes memory considerably easier to study in constrained closed drafting environments.

The Sushi Go Party environment only allows for a subset of total available unique cards to be used during a single game. In a typical game, 9 out of the 37 unique cards available will be selected for use. The game can become more competitive depending on the selection of cards chosen. We refer to this selection as a game configuration.


\paragraph{Memory} Strategies for playing closed drafting games incorporate an element of memorization. The hand a player has at the end of a round is passed to a neighbouring player, and in turn they receive a hand of the same size from a neighbouring player. This is circular, meaning that a player can use the knowledge of past hands in their decision-making. In our Sushi Go Party environment, we enable DRL agents to explicitly retain a memory input of past hands.

To interpret how explicit memory modulates agent behaviour, we propose two methods targeting distinct aspects of what it means to learn memory. The first is a classic t-test to compare whether agents given explicit memory and agents without the memory features have a statistically significant difference in game performance. This, however, does not mean that memory is helping in the way we think, especially if memory is not truly utilized by the model or if differences arise just because of the additional input dimensionality. We thus invent a new memorization test based on perturbing the memory portion of the input and measuring the KL divergence between the probability distributions over the action space:
$$\mathrm{MemInfluence}(\pi)=\E_{s'\sim\mathrm{Pert}(s)}[D_{KL}(\pi(\cdot|s')||\pi(\cdot|s))]$$
where $s'\sim\mathrm{Pert}(s)$ denotes changing one card in the memory of the previous player's hand. This metric more directly monitors how changing the memory changes the model's chosen action, regardless of how useful for end performance.

\paragraph{Generalization} To quantify the degree of similarity between game configurations, given $A,B$ as the sets of cards in play respectively, we define the set distance as: $$\mathrm{EnvSim}(A,B)=|(A\cup B)\setminus(A\cap B)|.$$
Then, we can describe the average generalization performance $k$ steps away as the expected performance of a model trained on an environment $A$ and evaluated on an environment $B$ with $\mathrm{EnvSim}$ of $k$. This a natural metric as we expect games with similar sets of cards in play to have similar game dynamics, and the more cards that are different, the more differences that arise. This is important to better understand and interpret the bounds of our agent's capabilities and their behavior out-of-distribution.

\paragraph{Interpretability} Other than understanding behaviors related to memory and generalizability, we would most like to simulate a faithful, context-dependent, and simple strategy to emulate the DRL agent. This is difficult in general, but perhaps more amenable in the closed drafting setting where we can easily construe hypothetical scenarios and obtain explicit preference rankings when players have just two cards left to choose from. In particular, to satisfy the aforementioned desiderata, we opt to use decision rules, an inherently interpretable method which maps short Boolean conjunctives of the inputs to the output classification. These Boolean conditions are easy to understand and explicitly optimize for the precision of these rules in explaining the model's actions (learning them involves fitting a tree ensemble and selecting the most precise yet disparate collection of rules from the decision tree branches). We use SkopeRules' implementation \cite{goix2020skope}, which generates if-then rules for each card which are filtered for precision and recall. These describe model preferences for one card over another, and we can sample datasets at particular situations to understand model strategy explaining context-specific behavior.

\section{Experiments}

\paragraph{Human Player Priority Lists}

We describe the preferences of different types of human players by constructing priority lists based on playing data from the website Board Game Arena. Data from 172,357 games of Sushi Go Party are collected. The average number of points derived from each card over all games is used to construct a priority list we call the ``Average Human Player Priority". It approximates the worth of each card to the average human player. Board Game Arena assigns an ELO score to each player registered on the website. Looking at the top six players with the highest ELO score and with at least 400 played games, we construct a priority list called the ``Elite Human Player Priority". These elite players have played a total of 7728 games.

\paragraph{Human-Like Agent}

To benchmark the performance of our DRL agents, we prepared a human-like agent which simply follows the ``Average Human Player Priority" list. Our trained DQN models will play against these human-like agents in the next experiments.

\paragraph{Model Training}

For all experiments, we train DQN models via self-play using the same neural network architecture (4 hidden layers of 128 units each). Other hyperparameters are hand-tuned until baseline performance. The reward function is the points scored plus 100 at the end on a win.

\paragraph{Trends by Game Configuration Distance}

To observe generalization dynamics of the DRL agents, we prepare 5 game configurations. One is based on the ``My First Meal" configuration found in the instruction manual for Sushi Go Party, intended for beginner players. Another is based on the ``Cutthroat Combo" configuration, intended for advanced players. We interpolate 3 game configurations in-between ``My First Meal" and ``Cutthroat Combo" by incrementally swapping one unique card for another. This gives us an incrementally more difficult series of game configurations. 

\begin{figure}[h]
    \centering
    \includegraphics[scale=0.45]{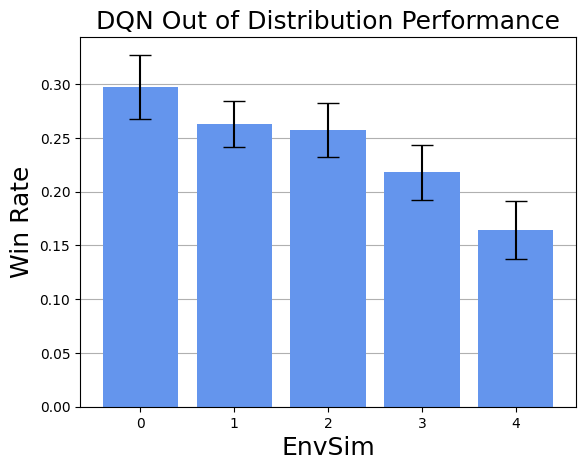}
    \caption{Win Rate Decreases the Larger the Set Distance}
    \label{fig:generalization}
\end{figure}

We train 10 DQN models for 10 epochs on each of the 5 game configurations, then test each trained model on all 5 game configurations against our human-like agent. There are 25 train-test configuration combinations, with an EnvSim value of between 0 and 4. Each of the 25 train-test configuration combinations was used to play 100 rounds of Sushi Go Party, 100 times. Each batch of 100 rounds used a different random seed. Win rate for the 100 rounds was calculated, and the average win rate was tallied and sorted by EnvSim (see Figure \ref{fig:generalization}).

\paragraph{Nuanced Memory Findings}


We use the ``My First Meal" configuration where we believe memory effects will be strong due to the presence of ``set completion" cards that need all elements to score any points, so their point values towards the end are highly dependent on remaining cards. We approximate our $\mathrm{MemInfluence}$ metric by sampling 10 random perturbations to a given state, and sampling 100 states from the second half of the round when the cards have rotated at least once around.

A t-test gives a statistically significant p-value of $1.4\times10^{-65}$, and the difference in mean reward is a substantial $14.08$. However, MemInfluence is $1.96\times 10^{-4}$, which means the distributions over next actions are essentially unchanged.

Thus, from a performance standpoint the DQN does significantly better with the memory explicitly given, intimating the failure of the model to capture memory effects on its own. However, we find that even the DQN trained with explicit memory does not appear to be influenced too heavily by it per the small MemInfluence metric.

One upside is that the MemInfluence metric finds the states where perturbing the memory causes the greatest change in chosen action. It turns out that this occurs in the second to last round, aligning with our intuition that this is when memory becomes especially important (for example, in knowing if you can finish a set), in which case the maximal difference from a single card change is a 6\% shift in the probability distribution for deciding between Tempura (a set completion card) and Soy Sauce (not a set completion card).


\paragraph{Interpretable Priority Comparisons}

\begin{figure}[H]
    \centering
    \includegraphics[scale=0.45]{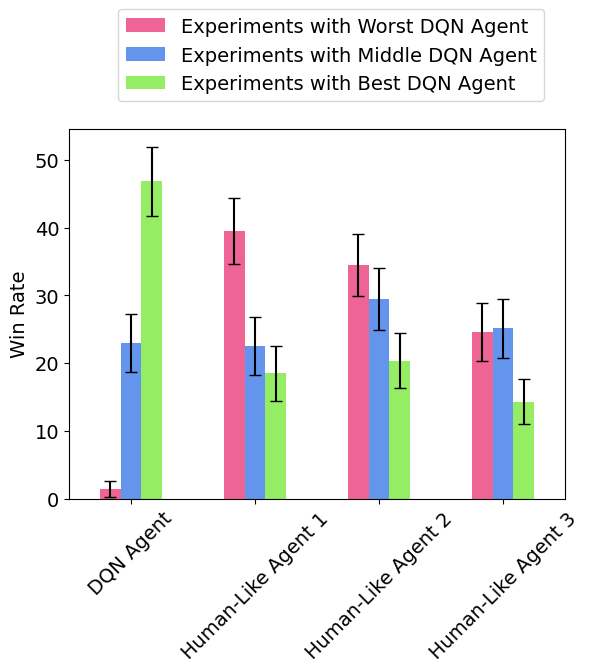}
    \caption{Worst, Medium, and Best DQN Agent Win Rate against Average Human-Like Agents; 500 Sets of 3 Rounds per Bar}
    \label{fig:three}
\end{figure}

\begin{figure*}[h]
    \centering
    \includegraphics[width=0.71\textwidth]{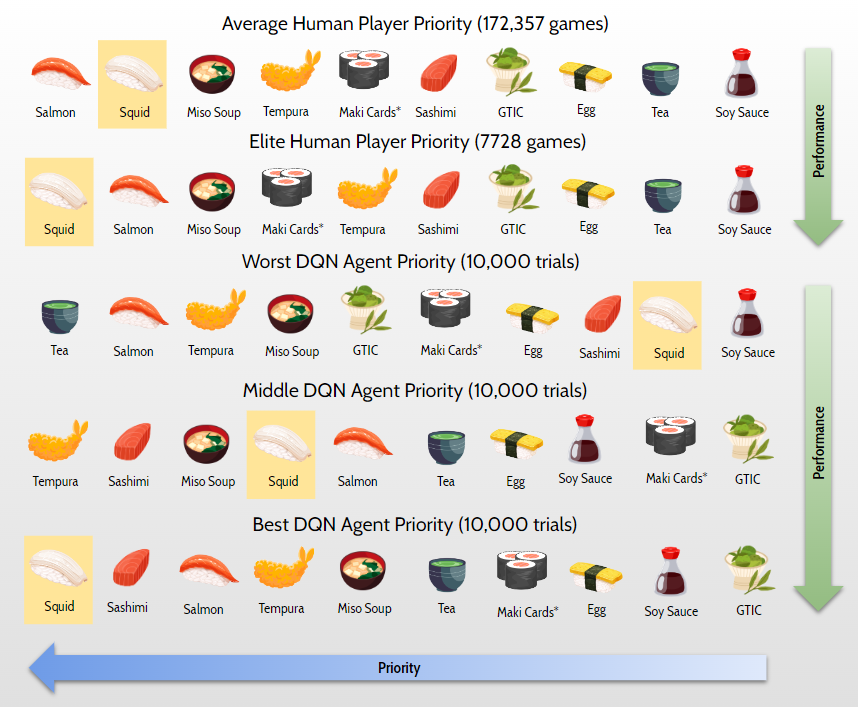}
    \caption[]{Human vs. DQN Priorities\footnotemark; Increased Preference for Squid as Performance Improves is Highlighted}
    \label{fig:interpretability}
\end{figure*}

We sample a dataset of observation-action pairs from the trajectory of three representative DQN agents with varying performances against human-like agents to fit decision rules on (see Figure \ref{fig:three}), looking at the second to last round with two cards remaining to get pairwise preferences. We use this to reconstruct the closest priority ranking for each of the three representative DQN agents. We compare these to the priority lists of the different types of human players (see Figure \ref{fig:interpretability}).


\footnotetext{There are several Maki cards in Sushi Go Party, but the Board Game Arena statistics and our analysis will consider them together.}

\section{Discussion}


We demonstrate that closed drafting is an understudied but highly useful class of environments for characterizing the interpretability, generalizability, and memory of DRL algorithms.
We use the unique properties of closed drafting (i.e., partial but easily learnable observability) to propose metrics for measuring the memory effects and the ability of DRL agents to learn memory. Additionally, the granularity of game configurations allows us to evaluate generalizability in terms of how out-of-distribution the test environment is. Beyond these implicit methods designed to better understand the impacts of specific memory and generalizability behaviors in simple ways, we construct explicit priority lists representing trained DRL agents that can easily be interpreted against the preferences of different types of human players to intuitively observe differences in DRL model decision making. We observe the movement of a single playable element (in this case Squid) through the priority list as explanation for the difference in performance between the models against the human-like agents.

Future work includes formalizing a general method of constructing priority lists and expanding this analysis to other closed drafting games. Running our experiments on more environments may shed light on more universal trends across closed drafting games. Simple enough closed drafting games could be fully solved with game theory and also accessible enough for a mechanistic interpretability probe to provide more evidence for the manifestation of memory.

\bibliography{aaai24}

\end{document}